\pdfoutput=1

\documentclass[11pt]{article}

\usepackage[preprint]{acl}

\usepackage{times}
\usepackage{latexsym}

\usepackage[T1]{fontenc}

\usepackage[utf8]{inputenc}

\usepackage{microtype}

\usepackage{inconsolata}

\usepackage{graphicx}
\newcommand{\cready}[1]{}

\title{The ShareLM Collection and Plugin: \\Contributing Human-Model Chats for the Benefit of the Community}

\author{
 \textbf{Shachar Don-Yehiya\textsuperscript{1}} \qquad
 \textbf{Leshem Choshen \textsuperscript{2,3}} \qquad
 \textbf{Omri Abend\textsuperscript{1}} \\
 \textsuperscript{1}The Hebrew University of Jerusalem,
 \textsuperscript{2}MIT,
 \textsuperscript{3}MIT-IBM Watson AI Lab \\
   \texttt{\{first.last\}@mail.huji.ac.il}
}

\begin{document}
\maketitle

\begin{abstract}
Human-model conversations provide a window into users' real-world scenarios, behavior, and needs, and thus are a valuable resource for model development and research. 
While for-profit companies collect user data through the APIs of their models, using it internally to improve their own models, the open source and research community lags behind. \cready{update citation in open human feedback paper}

We introduce the ShareLM collection, a unified set of human conversations with large language models, and its accompanying plugin, a Web extension for voluntarily contributing user-model conversations. Where few platforms share their chats, the ShareLM plugin adds this functionality, thus, allowing users to share conversations from most platforms. The plugin allows the user to rate their conversations, both at the conversation and the response levels, and delete conversations they prefer to keep private before they ever leave the user's local storage.
We release the plugin conversations as part of the ShareLM collection, and call for more community effort in the field of open human-model data.

The \textbf{code}, \textbf{plugin}, and \textbf{data} are available.\footnote{Code: \url{https://github.com/shachardon/share-lm}, Plugin: \url{https://chromewebstore.google.com/detail/sharelm-share-your-chat-c/nldoebkdaiidhceaphmipeclmlcbljmh}, Data: \url{https://huggingface.co/datasets/shachardon/ShareLM}}

\end{abstract}

\section{Introduction}

Recently, with the development of more capable models such as GPT4 \citep{openai2024gpt4} and LLAMA \citep{dubey2024llama3herdmodels}, interacting with large language models (LLMs) has become common not only among Machine Learning experts, but also the general public. Human users have natural language conversations with the models, and use them for a wide range of use cases \citep{zhao2024wildchat}. 
In turn, these conversations can be used for training and better-aligning models to human preferences, as they provide a window into the users' real-world scenarios and needs \citep{bai2022training}. The conversations are also important for other research aspects, such as cognitive and linguistic research revealing the gaps in the mode of interaction between models and humans \citep{donyehiya2023human}.

\begin{figure}[t]
\includegraphics[width=\columnwidth]{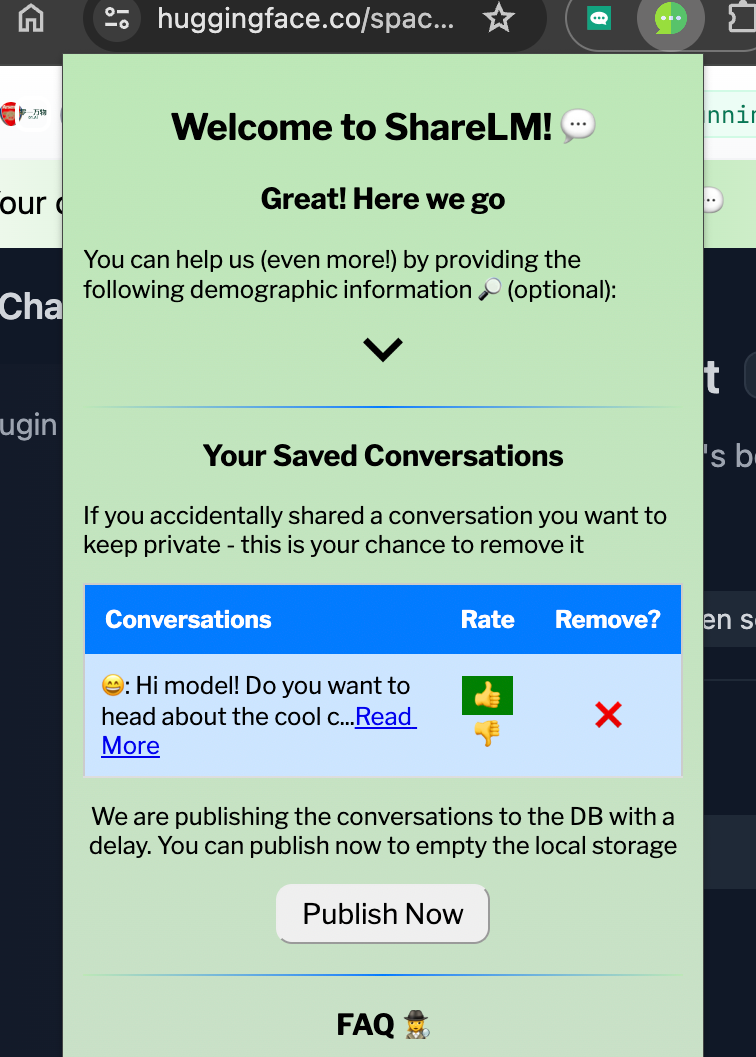}
\caption{The popup window. The user can go over their previous conversations from the last 24 hours and rate them or alternatively choose to delete them if they prefer to keep them private.}
\label{fig:fig_1}
\end{figure}

Despite being a cornerstone for LLM development and research, mechanisms for openly collecting and sharing human conversations and feedback are still underdeveloped. 
In the meantime, models developed by for-profit companies collect user-model conversations via their APIs to be used to further train their own models \citep{ouyang2022training}, leaving the open-source and research community far behind.
The development process of these ``closed models'' is not always transparent, and so are their data and data collection pipelines. These all make developing platforms and tools for collecting human-model conversations a high priority \citep{donyehiya2024futureopenhumanfeedback}. 


We collected existing human-model conversations datasets, and unified them under format. We call it the \textbf{ShareLM collection}.
Doing so, we recognized that most of the existing open datasets are treated as static collections rather than a living artifact that can dynamically grow (see \S\ref{sec:previous_work}). Unlike traditional Natural Language Processing datasets (e.g., grammatical error correction), human-model conversations and preferences vary across individuals and time \citep{pozzobon2023goodtrieveradaptivetoxicitymitigation}\cready{cite open human feedback}. Also, these types of data collection efforts are not something that private users can be part of and may lack in diversity \citep{pavlick2014language}. 

To overcome this, we introduce the \textbf{ShareLM plugin}, a Chrome extension that allows users to easily contribute their conversations with models. The ShareLM plugin collects the user's conversations with models, supporting multiple platforms and hence not limited to certain models, serving infrastructure or user interface. Among its main features, the plugin supports thumbs up/down rating, and a delayed upload that allows users to go over their conversations from the last 24 hours and remove those that they prefer to keep private before they ever leaved the user's local storage.
The plugin provides the end-point user with ownership of their data, allowing them to keep, delete and retrieve their data and to contribute it for the benefit of the community.
The plugin holds the potential to maintain an ever-growing dataset, up-to-date with users' conversations with the state-of-the-art models of the moment.


We release the conversations that are collected by the plugin as part of the broader ShareLM collection. We hope to see more efforts in the field and contributions to the ShareLM collection, with the aim of sharing open data.





\section{The ShareLM Collection} \label{sec:sharelm_collection}
We collected existing human-model conversations datasets that are publicly released.
As we focus on human-model conversations and realistic interactions, we exclude other conversation datasets such as human-human (such as in OpenAssistant \citep{kopf2024openassistant, zhang-etal-2018-personalizing}), model-model \citep{honovich-etal-2023-unnatural, wang2023selfinstruct} or human-model but not conversations \citep{nakano2021webgpt}. 

The current list of datasets contains the following; HH-RLHF \citep{bai2022training} which contains conversations of users with a closed model and their preferences, the dialog task of the bAbi-tasks \citep{weston2015towards}, the self-feeding chatbot data \citep{hancock-etal-2019-learning}, the Collective Cognition dataset (see \S\ref{sec:previous_work}), and PRISM \citep{kirk2024prism}, containing conversations and preferences of users born in 75 countries, residing in 38 countries with 21 different LLMs both opened and closed. Two more large datasets are WildChat \citep{zhao2024wildchat}, a dataset of over 1M conversations of users with ChatGPT, and the Chatbot Arena (see \S\ref{sec:previous_work}). The last two are gated datasets\footnote{\url{https://huggingface.co/docs/hub/datasets-gated}}, and thus require the user to conform to their terms of use prior to downloading them. 
We note that all these datasets were not collected by us originally and therefore we assume no responsibility. We ask the users to check each dataset directly for the appropriate citations and licenses. Still, those datasets mainly follow open licenses and we follow their licenses in the unification process.

Together with the conversations that were collected so far by the ShareLM plugin, the ShareLM collection currently contains over $2.3M$ conversations, from over $40$ different models. 

The unified format includes the following fields: \textit{conversation\_id} to identify each conversation, \textit{conversation} that contains the content of the conversation, \textit{model\_name} (if available), \textit{user\_id} an anonymized identifier of the user (if available), a \textit{timestamp} of the time the conversation was conducted (if available), the \textit{source} of the conversation i.e., from what dataset it was taken, \textit{user\_metadata} which contains demographic information of the user such as location (if available), and \textit{conversation\_metadata} that contains additional information regarding the conversation, e.g., language, user-feedback and more.


\section{Plugin Design and Architecture}
In the following section, we describe the design choices of the ShareLM plugin and the motivations behind them.
We start by outlining the leading principles, and then describe the implementation.  
    
\subsection{Main Principles}
Taking inspiration but more importantly lessons from the existing data collection platforms (see \S\ref{sec:previous_work}), we opt to design our plugin in accordance with the following principles:

\begin{enumerate}
    \item
        \textbf{Easy Usage.} The plugin should be 'transparent' to the user, i.e., its basic functionality should not require any extra effort from the user.
    \item 
        \textbf{Users own their data.} The plugin merely helps in sharing and providing an open license to the data that the user creates and owns.
    \item 
        \textbf{Enhanced User Control.} The user should be able to manage their data on their own, e.g., deleting unwanted conversations.
    \item 
        \textbf{Privacy.} The plugin must conform to established privacy standards.
    \item 
        \textbf{Inclusive Models List.} Our plugin should be a mediator for other platforms, potentially supporting every model out there. 
\end{enumerate}

These principles guided us through the plugin development, from the decision to implement it as a plugin, to the finer details such as the delayed upload feature.

\subsection{System Architecture}\label{sec:arch}

Upon installing the plugin and confirming the terms of use, the user is assigned a randomly generated user ID. We do not require the user to register and log in,
as we want to avoid unnecessary complications. 

The plugin works by identifying certain elements in the web page XML, according to the chat platform in use.
Currently, the plugin supports \textit{Gradio}\footnote{\url{https://www.gradio.app/}}, a web interface for various demos including chats, \textit{ChatUI}\footnote{\url{https://huggingface.co/docs/chat-ui/index}}, a web interface for chats, and ChatGPT. Those were chosen due to their popularity, e.g., Gradio and ChatUI are in frequent use in Huggingface Spaces\footnote{\url{https://huggingface.co/spaces}} and the ChatBot Arena (see \S\ref{sec:previous_work}). Nevertheless, adding support to new web platforms is easy\footnote{We were informed by private correspondence that external contributors are planning to extend the support to other interfaces}.

The plugin flow operates as follows: 
The user and model responses are periodically queried and collected, together with thumb-up/down notions if available. A check is performed to determine whether the current conversation is a new one or rather a continuation of the previous one.
Each new conversation is assigned a unique ID, a timestamp, and the current URL. The last is used to recognize what model the user was interacting with. The conversation is stored in a local database.

Upon a 24-hour delay, the conversations in the local database are posted to the server via a REST API, accompanied by the user ID and user/conversation metadata if available.

In turn, the server runs an anonymization script\footnote{\url{https://pypi.org/project/anonymization/}} on the conversation's content, to remove names, addresses, phone numbers, and more. We note that as part of the plugin terms of use, we ask users to avoid sharing conversations with such identifying details. The anonymization script is another line of protection, but no text shared should be assumed fully anonymous \citep{narayanan2008robust}. 
The server adds the new conversations to a PostgreSQL database.

Periodically, we release an updated version of the dataset\footnote{\url{https://huggingface.co/datasets/shachardon/ShareLM}}.
In the future, we plan to employ a fully automated release process, but for now, we validate it manually before uploading it for quality control. 

\begin{figure*}[t]
    \centering
    \includegraphics[width=1\linewidth]{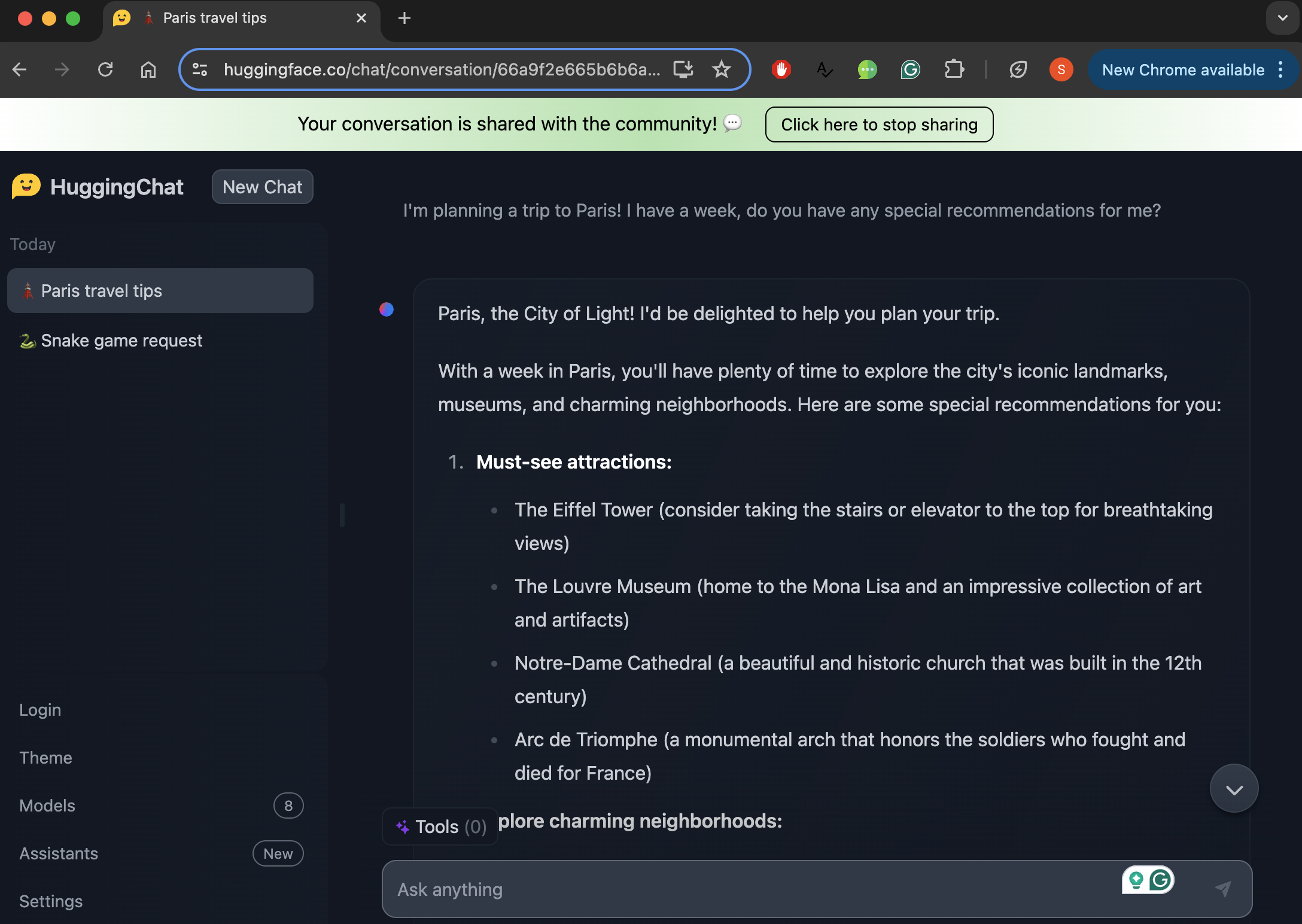}
    \caption{The recording banner is at the top of the window, indicating that the current chat demo (here ChatUI) is supported by the plugin and that the current conversation is recorded. Clicking on the "Click here to stop sharing" button will pause the conversation's recording.}
    \label{fig:recording_banner}
\end{figure*}

\begin{figure}[t]
\includegraphics[width=\columnwidth]{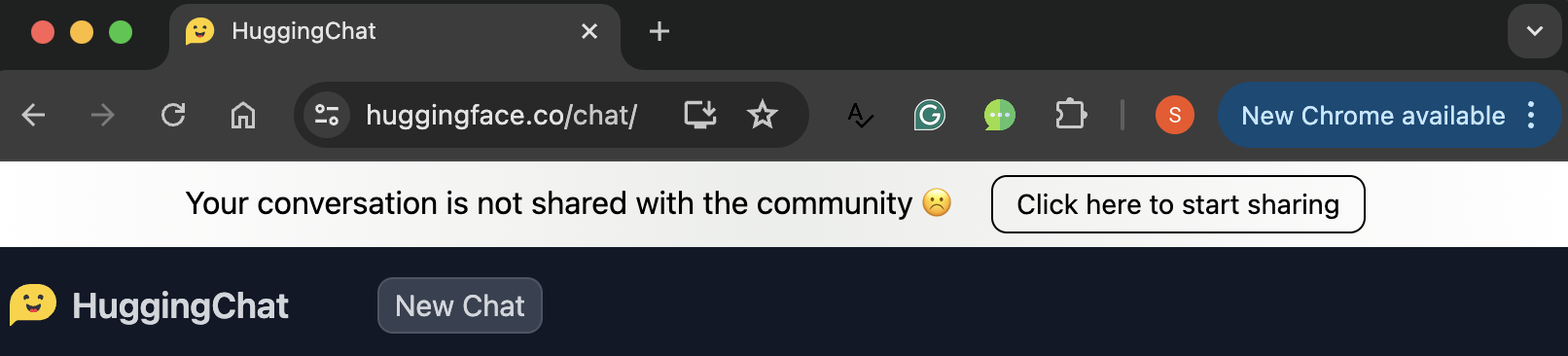}
\caption{The conversation collection is paused. Clicking on the "Click here to start sharing" button will start the conversation's recording.}
\label{fig:not_collecting_feedback}
\end{figure}

\section{The Plugin UI}

We describe the plugin UI components and usage.

\subsection{Terms of Use}

To activate the plugin after installation, the user needs to confirm the terms of use. The terms of use are available through the plugin popup (see \S\ref{sec:popup}), or the recording banner while in a supported demo (see next \S\ref{sec:banner}). We ask the users to avoid sharing conversations with identifying/sensitive content (names, e-mail addresses, etc.), as the content of the conversations will be publicly released. The full terms are available in the plugin repository and in App.~\S\ref{sec:terms}.

\subsection{The Recording Banner}\label{sec:banner}

The recording banner (see Fig.~\ref{fig:recording_banner}) is a thin strip at the top of the tab. The recording banner is available when the web page contains a supported demo interface (see \S\ref{sec:arch}). 
Seeing whether the current demo is supported is also possible through the extension icon. The icon is green when a supported interface exists, and gray otherwise.

The main role of the recording banner is to inform the user their conversations are recorded. In addition, it can be used to pause the conversation sharing. Clicking on the "Click Here to Stop Sharing" button will turn off the conversations collection (see Fig.~\ref{fig:not_collecting_feedback}). This is useful when conducting a conversation with identifying information that should be kept private.

With the recording banner, we balance between ease of use and control. We do not want to tire the user and require them to press buttons in order to record each conversation. On the other hand, we want the user to be aware that their conversations are recorded. Thus, the recording banner is designed to be visible but not interfere with normal use.

\subsection{The Popup}\label{sec:popup}

The plugin popup (see Fig.~\ref{fig:fig_1}) is where the more advanced features are concentrated. 

\paragraph{Demographic Details Form.}
Clicking on the down arrow at the top of the popup window opens a form of demographic details (Age, Country, and Gender). LLMs suffer from limited coverage of diverse human demographics in their training data, as their data usually comes from English speakers from narrow communities \citep{pavlick2014language}. Filling this form is voluntary, and can be of great help for studies focusing on diversity.

\paragraph{Saved Conversations Table.}
The saved conversations table contains all the user's recorded conversations from the last 24 hours. 
Clicking on a conversation extends it such that its full content is visible. The thumbs-up/down are used to rate the satisfaction of the user from the conversation as a whole. Rating the conversation and providing 'human feedback' is not mandatory, but it has great merit. Human feedback is a valuable resource for model development, as it allows better alignment of the model to human users' preferences.
Clicking on the red X button will delete the conversation from the local database, without it ever leaving the user storage.
Asking to delete past conversations through the contact form (available at the bottom of the popup) is always possible, but we note that after the dataset was already released it is very likely that someone has already downloaded and saved an old version of it.
Under the table, there is a ``Publish Now'' button that empties the local storage and publishes the conversations immediately.

\begin{figure}[t]
\includegraphics[width=\columnwidth]{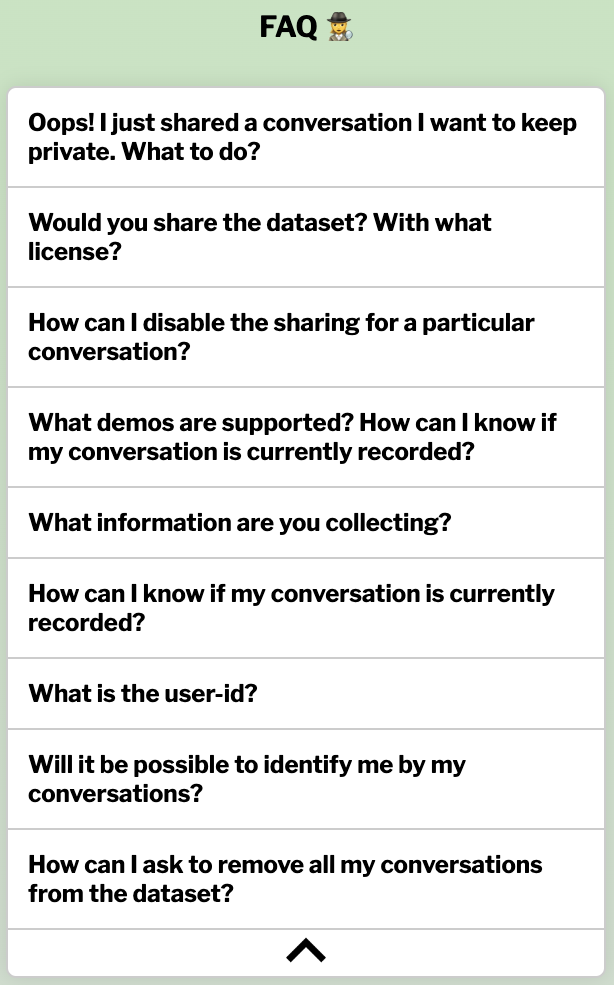}
\caption{The frequently asked questions section (in the popup window). Provides answers to common questions regarding the plugin.}
\label{fig:faq}
\end{figure}

\paragraph{Frequently Asked Questions.}
Under the conversations table, we include a frequently asked questions section, to answer common questions regarding the plugin (see Fig.~\ref{fig:faq}). There, we address questions regarding privacy (e.g., \textit{Will it be possible to identify me by my conversations?}), license (\textit{Would you share the dataset? With what license?}), ownership (\textit{How can I ask to remove all my conversations from the dataset?}) among others.

\paragraph{Contact Form.}
The contact form is used to request to remove already published conversations from the dataset. One can ask to remove their own conversation, or use the form to report others' conversations that violate the terms of use. When a user asks to remove their own conversations, they will be asked to include their user ID for identity verification. For that, they can use the 'Copy User ID' button which copies the user ID to their clipboard.

\begin{figure}[t]
\includegraphics[width=\columnwidth]{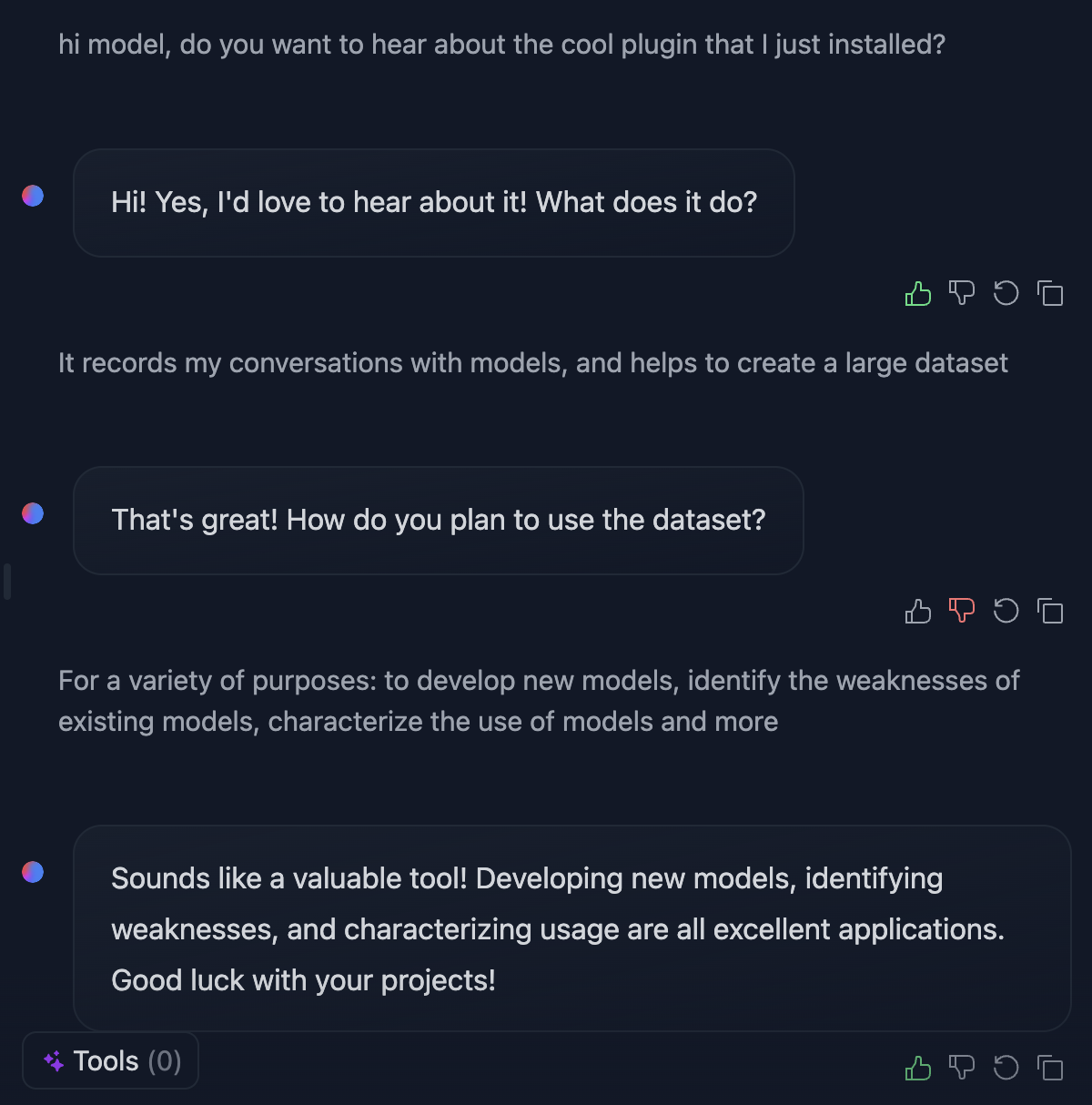}
\caption{Providing feedback through the chat interface. The user can rate each response separately, at the time of the interaction.}
\label{fig:providing_feedback}
\end{figure}

\section{Providing Human Feedback}

As was already mentioned in \S\ref{sec:popup}, in addition to collecting conversations the plugin can be used also for rating them. Providing feedback for a given conversation can be done in two manners. The first is through the plugin popup. As shown in Figure \ref{fig:fig_1}, after conducting the conversations, the user can mark their conversations with thumbs up/down to express their (dis)satisfaction with the entire conversation. The other way to provide feedback is through the chat interface in real-time, as demonstrated in Figure \ref{fig:providing_feedback}. The user can click the thumbs-up/down buttons separately for each model response. This allows better feedback granularity, and is also sometimes easier, as it does not require the user to go over their conversation again, but is instead done at the time of the interaction. We note that the per-response option is currently available for the ChatUI interface only.

\section{User Study}
We conducted a user study to evaluate the plugin. We asked $10$ participants to install and experiment with the plugin. 
On a scale of 1 (poor) to 5 (great), $9$ out of $10$ participants described the installation experience as $5$, and the average score was $4.8$. Some of them elaborated, saying that \textit{It was straight forwards, self explanatory} and \textit{Smooth sailing, really easy and nice}. The participants described the experience of using the plugin for the first time with an average score of $4.7$.
Half of the participants reported that they used the plugin popup to rate or delete some of their conversations. 
The participants described the UI with an average score of $4.7$, saying that it is \textit{Really responsive, quick, and neatly designed} and \textit{Easy to like a convo, to delete, and to understand the flow}. One of the participants said that \textit{Its refresh time is long}.
When asked how often do they use open models in their day-to-day activities on a scale of 1 (never) to 5 (all the time), the average score was $2.7$.

We discuss the last point in the limitations section (\S\ref{sec:limitations}).

\section{Previous Work}\label{sec:previous_work}

ShareGPT \footnote{\url{https://sharegpt.com/}}, a plugin for collecting and sharing conversations specifically with ChatGPT, is the closest to ours. Although not active these days, the ShareGPT plugin collected over 400,000 conversations and 90,000 of them were published as a dataset before its API was shut down.
Another effort is Collective Cognition\footnote{\url{https://huggingface.co/datasets/CollectiveCognition/chats-data-2023-10-16?row=11}}, a platform for collecting and tagging conversations with ChatGPT, also not active anymore. 
Unlike ShareGPT and Collective Cognition, our plugin is not limited to ChatGPT but rather focuses on open-source models. It is also easier to use and does not require the user to actively click buttons to share each conversation. 

The LMSYS's Chatbot Arena \citep{zheng2023judging} hosts various models, both open and closed, allowing users to access and interact with them in exchange for their conversations. Our plugin allows even more flexibility regarding the models in use, not limiting them to a closed list, and provides the user more control over their data.

The delayed upload and the ad-hock rating are not available on any of these platforms.   

Another line of work is the ``one-time collected'' datasets. These are not platforms for continuous data collection but rather high-quality datasets of human-model conversations that were crowdsourced (see \S\ref{sec:sharelm_collection}).
Although useful, these datasets are not updated over time, and hence can not solve the community needs alone \citep{pozzobon2023goodtrieveradaptivetoxicitymitigation, pavlick2014language}\cready{cite open human feedback}.

Argilla\footnote{\url{https://argilla.io}} is another open data platform, a collaboration tool for engineers and domain experts for dataset annotation. Unlike our plugin, it is used mostly for annotating existing datasets, not collecting new ones.

\section{Conclusions and Future Work}

We introduced the ShareLM Collection and Plugin, to support open human-model conversations and feedback. The code is openly available, and we welcome contributions. 
Although the number of users is still not large, the plugin already stimulates discussion among the community members, as well as external contributions (pull requests).


As we want to improve the user ownership experience, we plan to add a feature that allows the user to download their conversations that are currently still stored locally, as well as attribution and recognition metrics such as the number of conversations contributed so far.

Another future feature would be to recommend new models to users based on their popularity among other users.

Another line of future work would be to conduct research on model personalization, using the user ID to group all the user's conversations.

\section*{Limitations}\label{sec:limitations}

Collecting open human chats and feedback is a challenge. The ShareLM plugin tackles this from the end-point user's perspective, providing them with the ability to easily contribute their own conversations.
However, there are more places in the human-model interaction pipeline that can be used for contributing data. For example, the entity that serves the model can be responsible for collecting the conversations. This makes scaling easier, as we do not need each individual user to install a plugin.
On the other hand, the fact that the plugin is a mediator between the user and the serving platform, makes it more flexible, not limiting the contribution for certain platfroms or models.  


\section*{Ethics Statement} \label{sec:ethics}
The plugin and its use have been approved by the IRB of our institution.

\section*{Acknowledgments}
We thank Ben Burtenshaw for contributing the per-response thumbs up/down rating feature.

\bibliography{custom}

\appendix

\section{Terms of Use} \label{sec:terms}
Inspired by the release of the ChatGPT, the open-source community recently began to develop open access models with increased transparency about their development. The next challenge for democratizing large language models is data.

This extension collects the conversations you are having with open large language models (“chat-bots”). By using this extension, you are giving your permission to contribute your conversations’ content (both your side of the conversation, and the model’s) for creating an open-license chat-bot conversations dataset, a valuable resource for the open-source community. The conversations will be released with the most permissive license that is allowed by the specific model. This dataset will be a valuable resource for both model developers and researchers. Specifically, we plan to use this dataset to study and improve the nature of human- model interaction.

The extension supports a couple of chat-bots demos, mostly within Huggingface Spaces (https://huggingface.co/spaces). You will see a banner on the top of the demo page indicating it. You can choose not to share a particular conversation by clicking the ‘do not share’ button. As an additional precaution, the conversations are not posted to the database immediately. You can see the conversations from the last 24 hours in the extension popup window and remove them. To stop sharing your conversations permanently, please disable or remove the extension. Note that removing the extension does not delete the conversations you have already made.

Along with the conversation’s content, we are collecting the URL (to identify the model), GMT time and an anonymous user-id. Optionally, you can fill some demographic data (age, location, gender) and rate your satisfaction. We are not collecting any identifying metadata (such as IP address, local time, browser type, etc.). However, it is possible that you will be identified by the content of your conversations. Therefore, please avoid sharing conversations with Identifying/sensitive content (names, e-mail addresses, etc.), as the content of your conversations will be publicly released. If you accidentally shared the content of a conversation you prefer to keep private, please fill the contact form so we will remove it (available in the extension popup). You can ask to remove all your conversations at any time, but please note that after the dataset was already released it is very likely that someone has already downloaded and saved an old version of it. You are encouraged to use this form also for reporting conversations that are copyrighted, defamatory, threatening to others, violating of others' privacy, or that you view as harmful if released.

Please be advised that this extension is independently developed by us, and while we have put our best efforts into ensuring a smooth experience, it's important to note that there might be bugs or unforeseen issues. Your feedback is valuable to us, so please feel free to report any issues you may encounter.

The research is conducted by Shachar Don-Yehiya, Leshem Choshen and Omri Abend at the Hebrew University. 

For more questions, please contact us at shareLM.project@gmail.com.

Participation is from age 18 and over only.

Participation is voluntary. Thank you for your contribution!

\end{document}